\documentclass{article}

\usepackage[nonatbib,preprint]{nips_2018}

\usepackage[utf8]{inputenc} 
\usepackage[T1]{fontenc}    
\usepackage{hyperref}       
\usepackage{url}            
\usepackage{booktabs}       
\usepackage{amsfonts}       
\usepackage{nicefrac}       
\usepackage{microtype}      

\usepackage{researchpack}
\usepackage{wrapfig}
\newcommand{\methodname}{\textsc{caipi}}
\newcommand{\SelectQuery}{\ensuremath{\textsc{\textcolor{violet}{SelectQuery}}}}
\newcommand{\Fit}{\ensuremath{\textsc{\textcolor{violet}{Fit}}}}
\newcommand{\Explain}{\ensuremath{\textsc{\textcolor{blue}{Explain}}}}
\newcommand{\ToCounterExamples}{\ensuremath{\textsc{\textcolor{orange}{ToCounterExamples}}}}

\newcommand{\lime}{\textsc{lime}}
\newcommand{\rrr}{\textsc{rrr}}
\newcommand{\anchors}{\textsc{anchors}}

\usepackage{xcolor}

\title{``Why Should I Trust Interactive Learners?''\\
Explaining Interactive Queries of Classifiers to Users}

\author{
    Stefano Teso\\
    Department of Computer Science\\
    KU Leuven\\ Belgium\\
    \texttt{stefano.teso@cs.kuleuven.be}\\
    \And
    Kristian Kersting\\
    Department of Computer Science and \\
    Centre for Cognitive Science\\
    TU Darmstadt, Germany\\
    \texttt{kersting@cs.tu-darmstadt.de}
}

\begin{document}
\maketitle

\begin{abstract}
Although interactive learning puts the user into the loop, the learner remains mostly a black box for the user. Understanding the reasons behind queries and predictions is important when assessing how the learner works and, in turn, trust. Consequently, we propose the novel framework of {\it explanatory} interactive learning: in each step, the learner explains its interactive query to the user, and she
queries of any active classifier for visualizing explanations of the corresponding predictions.  We demonstrate that this can boost the predictive and explanatory powers of and the trust into the learned model, using text (e.g.~SVMs) and image classification (e.g.~neural networks) experiments as well as a user study.
\end{abstract}

\section{Introduction}

Trust lies at the foundation of major theories of interpersonal relationships in psychology~\cite{simpson2007psychological}. Building expectations through
interaction~\cite{kramer2001trust,mercier2011humans,chang2010seeing} and the
ability of interpreting the other's beliefs and
intentions~\cite{diyanni2012won} are necessary for (justifiably) establishing,
maintaining, and revoking trust. Hoffman {\it et al.}~\cite{hoffman2013trust} argue that
interpersonal trust depends on the ``perceived competence, benevolence
(or malevolence), understandability, and directability---the degree to which
the trustor can rapidly assert control or influence when something goes
wrong,'' and  Chang {\it et al.}~\cite{chang2010seeing} advocate that trust is ``dynamically updated based on experiences''.
Recent work shows that trust into machines
follows a similar pattern~\cite{hoffman2013trust,desai2013impact,waytz2014mind,wang2016trust}, with some
notable differences: it is often inappropriate to attribute
benevolence/malevolence to machines, and trust into machines suffers from different
biases than trust into individuals~\cite{hoffman2013trust}. These differences, however,
do not affect the argument that interaction and understandability are central
to trust in machine learners too. 
%
The
\emph{competence} of a classifier can be assessed by monitoring its behavior
and beliefs over time, \emph{directability} can be achieved by allowing the
user to actively teach the model how to act and what to believe, while
\emph{understandability} can be approached by explaining the model and its predictions.

Surprisingly, the link between interacting, explaining and building trust has been largely ignored by
the machine learning literature.
On one hand, existing machine learning explainers focus on the batch learning
setting only, and do not consider interaction between the user and the learner~\cite{bucilua2006model,ribeiro2016should,lundberg2016unexpected}.
Interactive learning frameworks such as active learning~\cite{settles2012active},
coactive learning~\cite{shivaswamy2015coactive}, and (to a lesser extent)
preference elicitation~\cite{pigozzi2016preferences}
do not consider the issue of
trust either.
In standard active learning, for instance, the model presents unlabelled
instances to a user, and in exchange obtains their label.  
This
interaction protocol is completely opaque: the user is oblivious to the
model's beliefs and reasons for predictions and to how they change in time, and
cannot see the consequences of her own instructions. In coactive learning, 
the user can correct the system, if necessary, but the predictions are not explained to her. 
So, why should users trust models learned interactively?

To fill this gap, we propose 
the novel framework of  
{\it explanatory interactive
learning}. Here the interaction takes the following form. In each step, the learner explains its interactive query to the user, and she
responds by correcting the prediction and explanations, if necessary, to provide feedback. We also present a model-agnostic method, called
\methodname{}, instantiating our framework for active learning.
\methodname{} extends active learning in several ways.
Akin to coactive learning~\cite{shivaswamy2015coactive}, query
instances are accompanied by the the model's corresponding \emph{predictions}.
This allows the user to check whether the model is right or wrong on
the chosen
instance. However, nothing prevents the model from being right (or wrong) for the wrong reasons, e.g., when there are ambiguities in the data such as
confounders~\cite{ross2017right}. To avoid this issue, \methodname{} accompanies predictions with
corresponding \emph{explanations},
computed by any local explainer of choice~\cite{ribeiro2016should,lundberg2016unexpected,ross2017right,ribeiro2018anchors};
in this paper, we advocate the use of \lime\footnote{This also explains the name \methodname, as \methodname{}rinhas are made out of \lime{}s.}~\cite{ribeiro2016should}, a simple model-agnostic
explainer that allows to easily compute explanations and present them to the
user as interpretable (visual) artifacts.
By witnessing the evolution of explanations---like a teacher supervising the
progress of a student---the user can see whether the model eventually ``gets
it''. 
Explanations can also improve the quality of feedback by focusing the
user's attention to parts or aspects of the instance deemed important by
the model~\cite{cakmak2014eliciting}.
Finally, the user can even correct the explanation presented to guide the learner.  This {\it correction} step is crucial for more
directly affecting the learner's beliefs and is integral to modulating
trust~\cite{hoffman2013trust,kulesza2015principles}.  Explanation corrections
also facilitate learning (the right concept), especially in problematic cases
that labels alone can not handle~\cite{ross2017right}, as shown by our
experiments.  Overall, \methodname\ is the first approach that employs explanation
corrections as an additional feedback channel in a model- and explainer-agnostic fashion.

To summarize, our main contributions are:
(1) explanatory interactive learning, an interactive learning framework
aiming at encouraging (or discouraging, if appropriate) trust into the model;
(2) a model- and explainer-agnostic implementation of the framework, called \methodname\, on top of active learning that makes
use of the \lime\ local explainer~\cite{ribeiro2016should};
(3) a simple data augmentation strategy for learning from explanation
corrections, and
(4) an empirical analysis showing that interacting through explanations can
modulate trust and improve learning effectiveness. 

We proceed as follows. First, we touch upon additional related work. Then we introduce 
explanatory interactive learning and derive \methodname. Before concluding, we present our empirical evaluation.

\section{Explainable, interactive, and trustworthy machine learning}
\label{sec:relatedwork}

There are two classes of explainable machine learners: global approaches aim to
explain a black-box model by converting it as a whole to a more interpretable
format~\cite{bucilua2006model,bastani2017interpreting}, while local approaches
interpret individual predictions~\cite{lundberg2016unexpected}.  Surprisingly,
they do not consider interaction between the user and the model.
On the other hand, existing approaches to interactive learning do not consider
the issue of explanations and trust.  This is true for active learning, coactive learning,
(active) imitation learning, \emph{etc.} In standard active learning, for
instance, the model presents unlabelled instances to a user, and in
exchange obtains their label.  This interaction protocol is opaque: the
user is oblivious to the model's beliefs and to how they change in time,
and can not see the consequences of her own instructions.
Given the centrality
of the user in recommendation, interactive preference elicitation approaches
make use of conversational interaction to improve trust and
directability~\cite{peintner2008preferences,chen2012critiquing}, but often rely
on rudimental learning strategies (if any).
Indeed, learning from explanations has been explored in concept
learning~\cite{mitchell1986explanation,dejong2011explanation} and probabilistc
logic programming~\cite{kimmig2007probabilistic}, where explanations are
themselves logical objects.  Unfortunately, these results are tied to
logic-based models and make use of rather opaque forms of explanations (e.g.
logic proofs), which can be difficult to grasp for non-experts. Explanatory
interactive learning instead leverages explanations for mainstream machine learning approaches.
More recently, researchers explored feature
supervision~\cite{raghavan2006active,raghavan2007interactive,druck2008learning,druck2009active,settles2011closing,attenberg2010unified}
and learning from
rationales~\cite{zaidan2007using,zaidan2008modeling,sharma2015active}, which
leverage both label- and feature-level (or sentence-level, for rationales)
supervision for improved learning efficiency. These works show that providing
rationales can be easy for human annotators~\cite{zaidan2007using}, sometimes even
more so than providing the labels themselves~\cite{raghavan2006active}. The
connection to directability and trust, however, is not explicitly made.
Explanatory interactive learning generalizes these ideas to arbitrary
classification tasks and models.  Actually, the techniques proposed in these
works are orthogonal to explanatory interactions and can be easily combined.
Finally, indeed, the UI community also investigated meaningful interaction strategies so that the user can build a mental model of the
system.  In~\cite{stumpf2009interacting}
the user is allowed to provide explanations, while~\cite{kulesza2015principles} provides
an explanation-centric approach to interactive teaching.  
These works however focus on simple machine learning models, like
Na\"ive Bayes, while explanatory interactive learning is much more general.

\section{Explanatory Interactive Learning}
\label{sec:background}


In {\it explanatory interactive learning}, a learner is able to interactively
query the user (or some other information source) to obtain the desired outputs
at data points. The interaction takes the following form.  At each step, the
learner considers a data point (labeled or unlabeled), predicts a label, and
provides explanations of its prediction. The user responds by correcting the
learner if necessary, providing a slightly improved---but not necessarily
optimal---feedback to the learner.

Let us now instantiate this schema to {\it explanatory active
learning}---combining active learning with local explainers. Indeed,
other interactive learning can be made explanatory too, including coactive learning
~\cite{shivaswamy2015coactive}, active imitation
learning~\cite{judah2012active}, and mixed-initiative interactive
learning~\cite{cakmak2011mixed}, but this is beyond the scope of this
paper.

{\bf Active learning.}  The active learning paradigm targets scenarios
where obtaining supervision has a non-negligible cost.  Here we cover the
basics of pool-based active learning, and refer the reader to two excellent
surveys~\cite{settles2012active,hanneke2014theory} for more details.  Let
$\calX$ be the space of instances and $\calY$ be the set of labels (e.g. $\calY
= \{\pm 1\}$).  Initially, the learner has access to a small set of labelled
examples $\calL \subseteq \calX \times \calY$ and a large pool of unlabelled
instances $\calU \subseteq \calX$.  The learner is allowed to query the label
of unlabelled instances (by paying a certain cost) to a user functioning as annotator, often a
human expert.  Once acquired, the labelled examples are added to $\calL$ and
used to update the model.  The overall goal is to maximize the model quality
while keeping the number of queries or the total cost at a minimum.  To this
end, the query instances are chosen to be as informative as possible, typically
by maximizing some informativeness criterion, such as the expected model
improvement~\cite{roy2001toward} or practical approximations thereof.  By
carefully selecting the instances to be labelled, active learning can enjoy
much better sample complexity than passive
learning~\cite{castro2006upper,balcan2010true}.  Prototypical active learners
include max-margin~\cite{tong2001support} and Bayesian
approaches~\cite{krause2007nonmyopic}; recently deep variants have
been proposed~\cite{gal2017deep}.

However, active---showing query data points---and even coactive learning---showing additionally the prediction of the query data point---
do not establish trust: informative selection
strategies just pick instances where the model is uncertain and likely wrong.  Thus, there
is a trade-off between query informativeness and user
``satisfaction,'' as noticed and explored in~\cite{schnabel2018short}.
In order to properly modulate trust into the model, we argue it is
essential to present explanations.

{\bf Local explainers.} 
There are two main strategies for
interpreting machine learning models.  Global approaches aim to explain the
model by converting it \emph{as a whole} to a more interpretable format
(e.g.~\cite{bucilua2006model,bastani2017interpreting}).  Local explainers---\lime{}~\cite{ribeiro2016should}, \rrr{}~\cite{ross2017right}, and
\anchors{}~\cite{ribeiro2018anchors}, among 
others--- instead focus on the arguably more approachable task of explaining
\emph{individual predictions}~\cite{lundberg2016unexpected}.  
While explainable
interactive learning can accommodate any local explainer, in our implementation
we use \lime{}, described next\footnote{Given a prediction, \rrr\ extracts an explanation from the input
gradients of the uninterpretable model.  It was shown that \lime\ and \rrr\
tend to capture similar information~\cite{ross2017right}.  The major difference
is that \rrr\ does not require sampling, but it is restricted to differentiable
models.  \anchors\ instead extracts high-precision rules from the target
prediction.}.
The 
idea of \lime{} (Local Interpretable Model-agnostic Explanations)
is simple: even though a classifier may rely on many uninterpretable features,
its decision surface around any given instance can be locally approximated by a
simple, interpretable \emph{local model}.  In \lime, the local model is defined
in terms of simple features encoding the presence or absence of \emph{basic
components}, such as 
words in a document or objects
in a picture\footnote{While not all problems admit explanations in terms of
elementary components, many of them do~\cite{ribeiro2016should}.}.
An explanation can be readily extracted from
such a model by reading off the contributions of the various components to the
target prediction and translating them to an interpretable (visual) artifact.
For instance, in document classification it makes sense to highlight the words
that support (or contradict) the target class.

More formally, let $f: \calX \to \calY$ be a classifier, for instance a Random
Forest or a Neural Network, $\hat{y} = f(x)$ the target prediction, and for
each basic component $i$ let $\psi_i(x)$ be the corresponding indicator
function. In order to explain the prediction, \lime\ produces an interpretable
model $g : \calX \to \calY$, based solely on the interpretable features
$\{\psi_i\}_i$, that approximates $f$ in the neighborhood of $x$.  Here $g$ can
be any sufficiently interpretable model, for instance a sparse linear
classifier or a shallow decision tree.  Computing $g$ amounts to solving
$\argmin_g \; \ell_x(f,g) + \Omega(g)$, where $\ell_x$ is a ``local loss'' that
measures the \emph{fidelity} of $g$ to $f$ in the neighborhood of $x$, and
$\Omega(g)$ is a regularization term that controls the complexity and
interpretability of $g$.

For the sake of simplicity, here we focus on \lime\ in
conjunction with sparse linear models of the form $g(x') =
\inner{\vw}{\vpsi(x)} + b$, where $\inner{\cdot}{\cdot}$ is the inner product.  In order to enhance interpretability, at most $k$
non-zero coefficients are allowed, where $k$ is sufficiently small (see
Section~\ref{sec:experiments} for the values we use).  Specifically, \lime\
measures the fidelity of the linear approximation with a ``local'' $L_2$
distance, namely $\ell_x(f,g) = \int_{x'} k(x,x') (f(x) - g(x'))^2 dx'$.  In
order to solve the optimization problem above, the integral is first
approximated by a sum over a large enough set $\calS \subseteq \calX$ of
instances sampled uniformly at random\footnote{In \lime{} the samples are taken
from the image of $\psi$, i.e., $\{ \psi(x) : x \in \calX \}$, and then mapped
back to $\calX$ to compute their predicted class. We omit this detail for
clarity.}.  Then, computing $g$ boils down to solving the sparsity-constrained
least-squares problem:
$
    g = \argmin\nolimits_{g} \; \sum\nolimits_{x' \in \, \calS} k(x,x') (f(x) - g(x'))^2 \quad \text{s.t.} \; \|\vw\|_0 \le k\;.
$
Note that $g$ does depend on both the target instance $x$ and on the prediction
$\hat{y} = f(x)$.  The relevance and polarity of all components can be readily
read off from the weights $\vw$: $|w_j| > 0$ suggests that the $j$th component
does contribute to the overall prediction, while $w_j > 0$ and $w_j < 0$ imply
that, when present, the $j$th component drives the prediction toward $\hat{y}$
or away from it, respectively.  Finally, this information is used to construct
a (visual) explanation~\cite{ribeiro2016should}.

{\bf Active learning with Local Interpretable Model-agnostic Explanations.}
%
Now, we have everything together for explanatory active learning and \methodname{}.
Specifically, we require black-box access to an active learner and an
explainer.  We assume that the active learner provides a procedure
$\SelectQuery(f, \calU)$ for selecting an informative instance $x \in \calU$
based on the current model $f$, and a procedure $\Fit(\calL)$ for fitting a new
model (or update the current model) on the examples in $\calL$.  The explainer
is assumed to provide a procedure $\Explain(f, x, \hat{y})$ for explaining a
particular prediction $\hat{y} = f(x)$.  The framework is intended to work for
any reasonable learner and explainer.
Here we employ \lime{} (described above) for computing an interpretable model
locally around the queries in order to visualize explanations for current
predictions.

The pseudo-code of \methodname\ is listed in Alg.~\ref{alg:framework} and
follows the standard active learning loop.  At each iteration $t = 1,
\ldots, T$ an instance $x \in \calU$ is chosen using the query selection
strategy implemented by the \SelectQuery\ procedure.  Then its label $\hat{y}$
is predicted using the current model $f$, and \Explain\ is used to produce an
explanation $\hat{z}$ of the prediction.  The triple $(x, \hat{y}, \hat{z})$ is
presented to the user as a (visual) artifact.  The user checks the
prediction and the explanation for correctness, and provides the required
feedback. Upon receiving the feedback, the
system updates $\calU$ and $\calL$ accordingly and re-fits the model.  The loop
terminates when the iteration budget $T$ is reached or the model is good
enough. 

During interaction between the system and the user, three cases can occur:
\begin{algorithm}[tb]
    \begin{algorithmic}[1]
            \State $f \gets \Fit(\calL)$
            \Repeat
                \State $x \gets \SelectQuery(f, \calU)$
                \State $\hat{y} \gets f(x)$
                \State $\hat{z} \gets \Explain(f, x, \hat{y})$
                \State Present $x$, $\hat{y}$, and $\hat{z}$ to the user, obtain $y$ and explanation correction $\calC$
                \State $\{(\bar{x}_1, \bar{y}_1), \ldots, (\bar{x}_c, \bar{y}_c)\} \gets \ToCounterExamples(\calC)$
                \State $\calL \gets \calL \cup \{(x, y)\} \cup \{(\bar{x}_i, \bar{y}_i)\}_{i=1}^c, \; \calU \gets \calU \setminus (\{x\} \cup \{\bar{x}_i\}_{i=1}^c)$ \label{eq:updatesets}
                \State $f \gets \Fit(\calL)$
            \Until{budget $T$ is exhausted or $f$ is good enough}
            \State $\textbf{return} f$
    \end{algorithmic}
    \caption{\label{alg:framework} The \methodname\ algorithm takes as input: $\calL$ is the
    set of labelled examples, $\calU$ is the set of unlabelled instances, and
    $T$ is the iteration budget.}
\end{algorithm}
%
    {\bf (1) Right for the right reasons:} The prediction and the explanation are both correct.  In this case,
        no feedback is requested.
    {\bf (2)Wrong for the wrong reasons:}  The prediction is wrong.  As in active learning, we ask the user
        to provide the correct label. Indeed, the explanation is also
        necessarily wrong, but we currently do not require the user to act on it.
    {\bf (3) Right for the wrong reasons:}  The prediction is correct but the explanation is wrong.  We ask the
        user to provide an explanation \emph{correction} $\calC$.


The ``right for the wrong reasons'' case is novel in active learning, and we propose {\it explanation corrections} to deal with it. They can assume different meanings
depending on whether the focus is on component relevance, polarity, or relative
importance (ranking), among others.  In our experiments we ask the annotator to
indicate the components that have been wrongly identified by the explanation as
relevant, that is,
$\calC = \{j : |w_j| > 0 \land \text{the user believes the $j$th component to be irrelevant}\}$.
In document classification, $\calC$ would be the set of words that are
irrelevant according to the user, but relevant for the model's explanation.

Given the correction $\calC$, we are faced with the problem of explaining it
back to the learner.  We propose a simple strategy to achieve this.
This strategy is embodied by the \ToCounterExamples\ procedure, which converts
$\calC$ to a set of \emph{counterexamples}.  These aim at teaching the learner
not to depend on the irrelevant components.  In particular, for every $j \in
\calC$ we generate $c$ examples $(\bar{x}_1, \bar{y}_1), \ldots, (\bar{x}_c,
\bar{y}_c)$, where $c$ is an application-specific constant.  Here, the labels
$\bar{y}_i$ are identical to the prediction $\hat{y}$.  The instances
$\bar{x}_i$, $i = 1, \ldots, c$ are also identical to the query $x$, except
that the $j$th component (i.e. $\psi_j(x)$) has been either randomized, changed to an alternative
value, or substituted with the value of the $j$th component appearing in other
training examples of the same class.  In sudoku, each $\bar{x}_i$ would be a
copy of the query sudoku $x$ where the cells in $\calC$ have been (for
instance) filled with random numbers consistent with the predicted label.  This
process produces $c |\calC|$ counterexamples\footnote{Instead of instantiating
\emph{all} possible counterexamples, it may be more efficient to only
instantiate the ones that influence the current model, i.e.,
\emph{adversarial} ones.  This is an interesting avenue for future work.}, which are added to $\calL$.

This data augmentation procedure is model-agnostic, but alternatives do exist,
for instance contrastive examples~\cite{zaidan2007using} and feature
ranking~\cite{small2011constrained} for SVMs and constraints on the input
gradients for differentiable models~\cite{ross2017right}.  These may be more
effective in practice, and indeed \methodname\ can accomodate all of them.
However, since our strategy is both model- and explainer-agnostic, in the
remainder we will stick to it for maximum generality.

{\bf Cognitive cost and confusion.}
Good explanations can effectively reduce the effort required to understand the
query and facilitate answering it.  Furthermore, by focusing the user's
attention to components relevant \emph{for the model}, in the same spirit of
teaching guidance~\cite{cakmak2014eliciting}, explanations can improve the
quality of the obtained feedback.  In some cases, however, explanations can be
problematic.
%
For instance, at the beginning of the learning process, the model is likely to
underfit and its explanations can be arbitrarily bad.  While these provide an
opportunity for gathering informative corrections, they can also confuse the
user by focusing her attention on irrelevant aspects of the query, but we note that the
same holds when no explanation is shown at all, since the user may fail to
notice the truly relevant components.
The degree of this effect depends
on how much the user relies on the explanation.  In practice, it makes sense to
only enable explanations after a number of burn-in iterations.


{\bf Mathematical intuition.} To gain some understanding of \methodname, let us
consider the case of linear max-margin classifiers.  Let $f(x) =
\inner{\vw}{\vphi(x)} + b$ be a linear classifier over two features, $\phi_1$
and $\phi_2$, of which only the first is relevant.  Fig.~\ref{fig:toy} (left)
shows that $f(x)$ (red line) uses $\phi_2$ to correcy classify a negative
example $x_i$.  In order to obtain a better model (e.g. the green line), the
simplest solution would be to enforce an orthogonality constraint
$\inner{\vw}{[0, 1]} = 0$ during learning.
Counterexamples follow the same principle.  In the separable case, the
counterexamples $\{\bar{x}_{i\ell}\}_{\ell=1}^c$ amount to additional
max-margin constraints~\cite{cortes1995support} of the form $y_i
\inner{\vw}{\vphi(\bar{x}_{i\ell})} \ge 1$.  The only ones that influence the
model are those on the margin, for which strict equality holds.  For all pairs
of such counterexamples $\ell, \ell'$ it holds that
$\inner{\vw}{\vphi(\bar{x}_{i\ell})} = \inner{\vw}{\vphi(\bar{x}_{i\ell'})}$,
or equivalently $\inner{\vw}{\vdelta_{i\ell} - \vdelta_{i\ell'}} = 0$,
where $\vdelta_{i\ell} = \vphi(\bar{x}_{i\ell}) - \vphi(x_i)$.  In other words,
the counterexamples encurage orthogonality between $\vw$ and the correction
vectors $\vdelta_{i\ell} - \vdelta_{i\ell'}$, thus approximating the
orthogonality constraint above.

\section{Empirical analysis}
\label{sec:experiments}

Our intention here is to address empirically the following questions:
(\textbf{RQ1}) Can explanations (and their consistency over time) appropriately
modulate the user's trust into the model?
(\textbf{RQ2}) Can explanation corrections lead to better models?
(\textbf{RQ3}) Do the explanations necessarily improve as the learner obtains
more labels?
(\textbf{RQ4}) Does the magnitude of this effect depend on the specific
learner?

{\bf (RQ1) User study.}  
We designed a questionnaire about a machine that learns a simple concept by
querying labels (but \emph{not} explanation corrections) to an annotator.  The
questionnaire was administered to 17 randomly selected undergraduate students
from an introductory course on deep learning.
\begin{figure}[tb]
    \centering
    \begin{tabular}{cc}
        \includegraphics[height=0.18\textwidth]{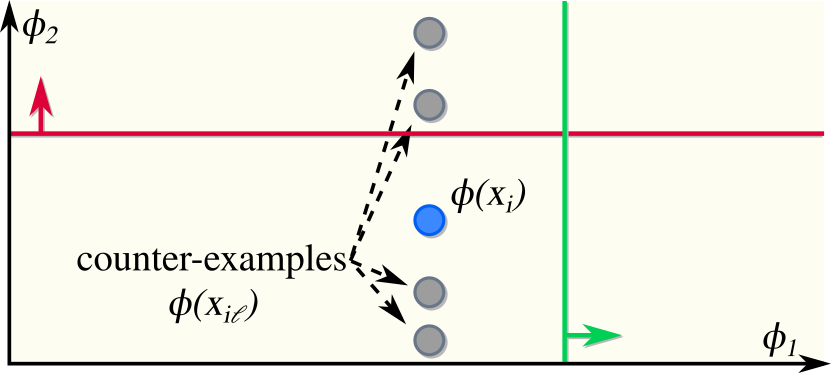}
        &
        \includegraphics[height=0.18\textwidth]{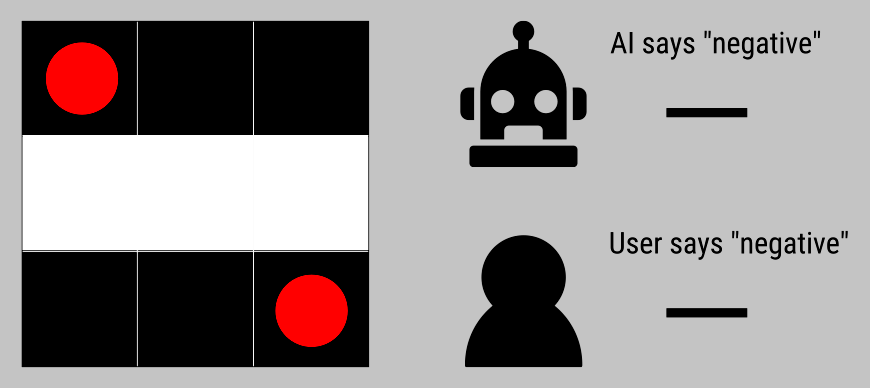}
    \end{tabular}
    \caption{
        \label{fig:toy}
        Left: mathematical intuition for the counterexample strategy.  Right: example
        training rounds as presented in the questionnaire.  The classification
        is correct but the explanation shows that the two most relevant pixels
        do not match the correct classification rule (as in S3). (Best viewed
        in color).
    }
\end{figure}
\begin{table}[b]
\centering
    \begin{tabular}{r|c|c|c}
                            & \textbf{Q1} & \textbf{Q2}   & \textbf{Q3}
        \\
        \hline
        \textbf{S1}  & 64.7\%      & 35.3\%        & 82.4\%
        \\
        \hline
        \textbf{S2}  & 76.5\%      & 64.7\%        & 70.6\%
        \\
        \hline
        \textbf{S3}  & 29.4\%      & 11.8\%        & 41.2\%
        \\
    \end{tabular} \quad \quad
     \begin{tabular}{l|c|ccc|c}
                & No corr.  & $c=1$ & $c=3$  & $c=5$ & IG \\
        \hline
          Train   & {\bf 0.978}     & 0.938 &  0.922 &  0.924 & 0.898 \\
        Test    & 0.482     & 0.821 & 0.851 &  {\bf 0.858} & 0.853 
    \end{tabular}
    \caption{ (Left-hand side table) \label{fig:answers} User study: percentage of ``yes'' answers. (Right-hand side table) \label{fig:fashion} Accuracy on the fashion MNIST dataset of an
    MLP without corrections (left), with our counterexample corrections using
    varying $c$ (middle), and with input gradient constraints~\cite{ross2017right}
    (right).}
\end{table}
We designed a toy binary classification problem (inspired
by~\cite{ross2017right}) about classifying small ($3 \times 3$) black-and-white
images.  The subjects were told that an image is positive if the two top corners
are white and negative otherwise.  Then they were shown three learning sessions
consisting of five query/feedback rounds each.
In session 1 (\textbf{S1}) every round included the images chosen by the model,
the corresponding prediction, and the label provided by a knowledgeable
annotator.  No explanations were shown.  
%
%
The predictions are
wrong for the first three rounds and correct in the last two.  Sessions 2 and 3
(\textbf{S2}, \textbf{S3}) were identical to S1, meaning that at every round
\emph{the same example, prediction and feedback label} were shown, but now
explanations were also provided.  The explanations highlighted the two most
relevant pixels, as in Fig.~\ref{fig:toy} (right).  In S2 the explanations
converged to the correct rule---they highlight the two top corners---from the
fourth round onwards, while in S3 they did not.  Removing the explanations
reduces both S2 and S3 to S1.
After each session, the subjects were asked three questions:
(\textbf{Q1}) ``Do you believe that the AI system eventually learned to
classify images correctly?''
(\textbf{Q2}) ``Do you believe that the AI system eventually learned the
correct classification rule?''
(\textbf{Q3}) ``Would you like to further assess the AI system by checking
whether it classifies 10 random images correctly?''
The first two questions test the subject's uncertainty in the predictive
ability and beliefs of the classifier, respectively, while the last one tests
the relationship between predictive accuracy (but \emph{not} explanation
correctness) and expected uncertainty reduction.
The percentage of ``yes'' answers  is reported in Tab.~\ref{fig:answers} (left).

As expected, the uncertainty in the model's correctness depends heavily on what information
channels are enabled.  When no explanations are shown (S1), only
35\% of the subjects assert to believe that the model learned the correct rule (Q2).
This percentage almost doubles (65\%) when explanations are shown and converge
to the correct rule (S2).  The need to see more examples also lowers from 82\%
to 71\%, but does not drop to zero.  This 
reflects the fact that five
rounds are not enough to reduce the subject's uncertainty to low enough levels.
The percentage of subjects asserting that the classifier produces correct
predictions (regardless of the learned rule, Q1) also increases from 65\% to 77\%
when correct explanations are shown (S2).  When the explanations do not converge
(S3), the trend is reversed: Q1 drops to 29\% and Q2 to 12\%, that is, most
subjects do not believe that the model's behavior and beliefs are in any way
correct.  This is the only setting where Q3 drops below 50\% (41\%): witnessing
that the model's beliefs do not match the target rule induces distrust (with
high certainty).  This 
confirms the previous finding that trust into
machines drops when wrong behavior is witnessed~\cite{hoffman2013trust}.  We
can therefore answer \textbf{RQ1} affirmatively: augmenting interaction with
explanations does appropriately drive trust into the model.


Next to the user study, we considered simulated users---as it is common for active learning--- to investigate
{\bf (RQ2--4)}. 
To this aim, we implemented \methodname\ on top of several
standard active learners and applied it to different learning tasks.  Note that
our goal here is to evaluate the contribution of explanation feedback, not the
learners themselves.  Indeed, \methodname\ can trivially accommodate more
advanced models than the ones employed here.  In all cases, the model's
explanations are computed with \lime~\footnote{Being based on sampling, \lime\
can sometimes output different explanations for the same prediction.  We
substantially improve its stability by running it 10 times and keeping the $k$
components identified most often.}.  As is common in active learning, we
simulate a human annotator that provides correct labels.  Explanation
corrections are also assumed to be correct and complete (i.e. they identify all
false positive components), for simplicity\footnote{In practice corrections may
be incomplete or noisy, especially when dealing with non-experts.  This can be
handled by, e.g., down-weighting the counterexamples.}.  The specifics of the
correction strategy are described in the next paragraphs.  Our experimental
setup is available at: URL ANONYMIZED FOR REVIEWING.

{\bf (RQ2) Evaluation on a passive setting.}  
We applied
our data augmentation strategy to a decoy variant of fashion MNIST, a fashion
product recognition dataset\footnote{From
\texttt{https://github.com/zalandoresearch/fashion-mnist}}.  
The dataset
includes 70,000 images over 10 classes.  All images were corrupted by
introducing confounders, that is, $4 \times 4$ patches of pixels in randomly
chosen corners whose shade is a function of the label in the training set and
random in the test set (see~\cite{ross2017right} for details).
The average test set accuracy of a multilayer perceptron (with the same
hyperparameters as in~\cite{ross2017right}) is reported in Tab.~\ref{fig:fashion} (right) for three correction
strategies: no corrections, our counterexample strategy (CE), and the
input-gradient constraints proposed by~\cite{ross2017right} (IG).  For CE, for
every training image we added $c = 1, 3, 5$ counterexamples where the
decoy pixels are randomized.  When no corrections are given, the accuracy on
the test set is $48\%$: the confounders completely fool the network.  Providing
even a single counterexample increases the accuracy to $82\%$, i.e., the
effect of confounders drops drastically.  With more counterexamples the
accuracy passes the one of IG ($85\%$).  This shows that (\textbf{RQ2})
counterexamples---and therefore explanation corrections---are an effective
measure for improving the model in terms of both predictive performance and
beliefs.

{\bf (RQ3,4) Actively choosing among concepts.}
We applied \methodname\ to the ``colors'' dataset
of~\cite{ross2017right}.
The goal is to
classify $5 \times 5$ images with four possible colors.  An image is positive
if either the four corner pixels have the same color (rule 0) or the three top
middle pixels have different colors (rule 1).  Crucially, the dataset only
includes images where either both rules hold or neither does, that is, labels
alone can not disambiguate between the two rules.  Explanations highlight the
$k$ most relevant pixels, and corrections indicate the pixels that are wrongly
identified as relevant.  In the counterexamples, the wrongly identified pixels
are recolored using all possible alternative colors consistent with $\hat{y}$
\footnote{In all experiments we always discard counterexamples that appear in
the test set, for correctness.}.
The features are of the form ``pixel $i$ has the same color as pixel $j$'' for
all $i, j = 1, \ldots, 25$, $i < j$.  In this space, the rules can be represented
by sparse hyperplanes.
We select each rule in turn and
provide corrections according to it, and then check whether the feedback drives
the classifier toward
it. $k$ was set to $4$ for rule 0 and to
$3$ for rule 1.  All measurements are 10-fold cross-validated.

In a first step, we considered a standard $L_2$ SVM active learner with the
closest-to-margin query selection heuristic~\cite{settles2012active}.  This
classifier can in principle represent both rules, but it is not suited for
learning sparse concepts.  Indeed, the SVM struggles to learn both rules, and
the counterexamples have little effect on it (see the Appendix
for the complete results).  This is plausible since the $L_2$ norm cannot
capture the underlying sparse concept: even though corrections try to drive the
model toward it, the $L_2$ SVM can still learn \emph{both} rules (as shown by
the coefficient curves) without a problem.  In other words, the model is not
constrained enough.

\begin{figure}[tb]
    \centering
    \begin{tabular}{cc|cc}
        \includegraphics[width=0.22\textwidth]{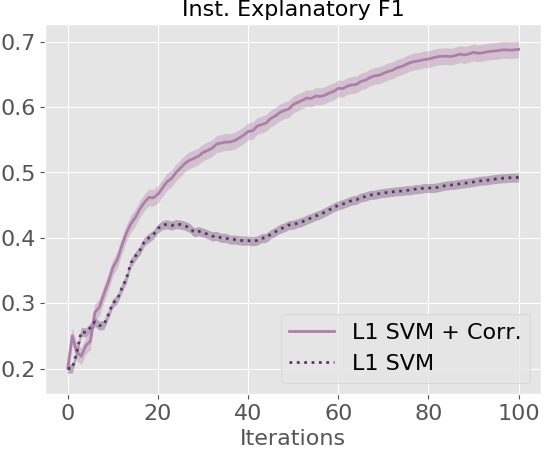}
        &
        \includegraphics[width=0.22\textwidth]{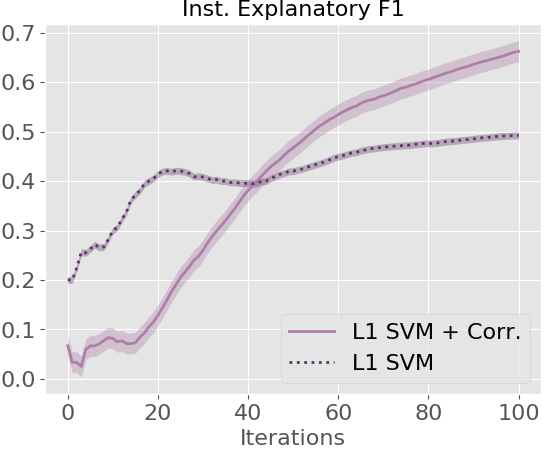}
        &
        \includegraphics[width=0.22\textwidth]{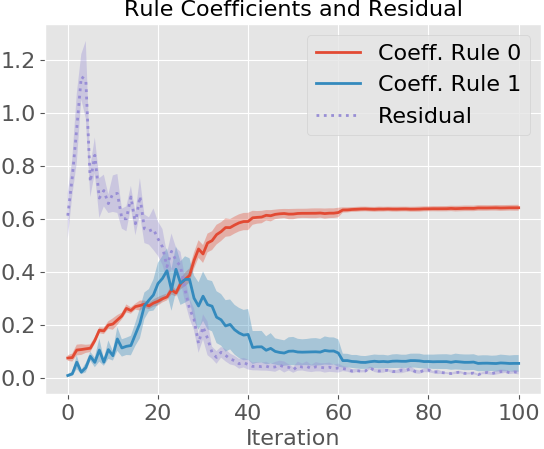}
        &
        \includegraphics[width=0.22\textwidth]{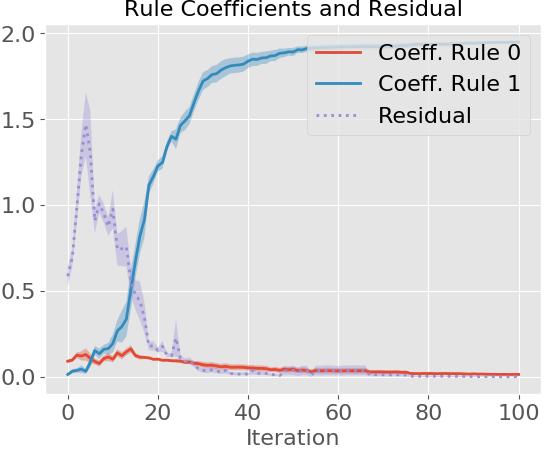}
    \end{tabular}
    \caption{\label{fig:colorsl1} $L_1$ SVM on the colors problem.  Left:
    instantaneous $F_1$ score of the \lime{} explanations
    for rule 0 (leftmost) and rule 1 (left middle).
    Right: decomposition of the learned weight vector when the corrections push
    toward rule 0 (right middle) and rule 1 (rightmost).  (Best viewed in
    color)}
\end{figure}

An $L_1$ SVM, an active learner tailored for sparse
concepts~\cite{zhu20041norm}, fares much better.  Our results show that the
rules greatly benefit this model.  To evaluate their effect, we compute the
average instantaneous $F_1$ score of the pixels identified by \lime{} w.r.t.
the pixels truly relevant for the selected rule.  This measures the quality of
the explanations presented to the user.  In addition, we measure the objective
quality of the model by decomposing the learned weights using least-squares as
$\vw = \alpha_0 \vw^*_0 + \alpha_1 \vw^*_1 + \text{residual}$, where $\vw^*_i$
is the ``perfect'' weight vector of rule $i = 0, 1$.  The instantaneous $F_1$
and change in coefficients can be viewed in Fig.~\ref{fig:colorsl1}.
Now that the model can capture the
target concepts, the contribution of counterexamples is very noticeable: the
$L_1$ SVM is biased toward rule 1, as it is sparser (data not shown), but it
veers clearly toward rule 0 when corrections are provided and learns rule 1
faster when corrections push toward it.  These results show clearly that
explanation feedback can drive the classifier toward the right concept, so long
as the chosen model can capture it clearly.

{\bf (RQ3,4) Active learning for document classification.}  Finally, we applied
\methodname{} to distinguishing 
between ``Atheism'' and ``Christian'' posts in
the 20 newsgroups dataset \footnote{From:
\texttt{http://kdd.ics.uci.edu/databases/20newsgroups/20newsgroups.data.html}}
using logistic regression with uncertainty sampling. Headers and footers were
removed; only
\begin{wrapfigure}{r}{22em}
   \centering
   \begin{tabular}{cc}
       \includegraphics[width=0.25\textwidth]{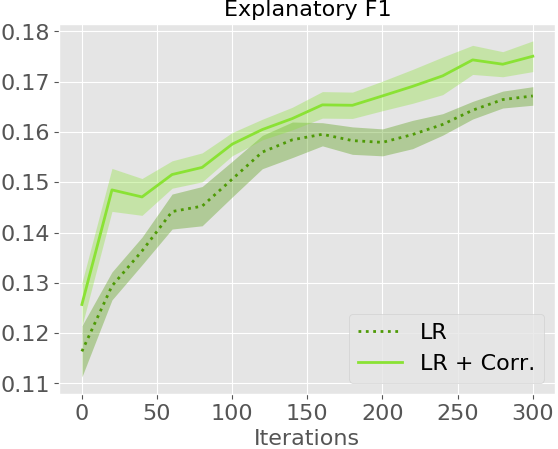}
       &
       \includegraphics[width=0.25\textwidth]{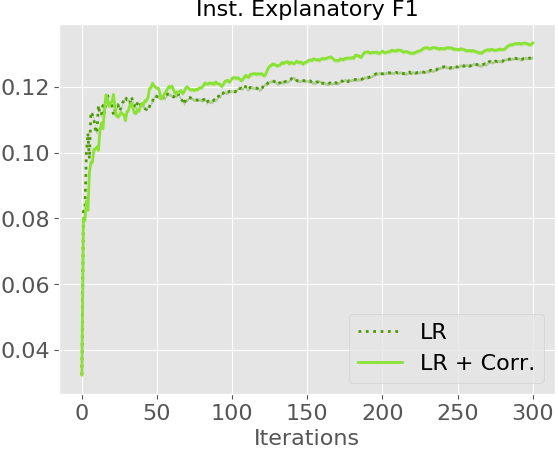}
   \end{tabular}
   \caption{\label{fig:newsgroups} Logistic regression on 20 newsgroups. (Best viewed in color)}
\end{wrapfigure}
 adjectives, adverbs, nouns, and verbs were kept and stemmed.  As
gold standard for the explanations, we selected $\approx\!\frac{1}{5}$ of the
words as relevant using feature selection.  Here the \lime{}-provided
explanations identified the $k$ most relevant words, while corrections
identified the falsely relevant words.  For each document, $k$ was set to
the number of truly relevant words.  To showcase \methodname{}'s flexibility,
the counterexamples were generated with the strategy proposed
in~\cite{zaidan2007using}, adapted to produce feedback based on the
falsely relevant words only.
The $10$-fold cross-validated results can be found in
Fig.~\ref{fig:newsgroups}.  The plots show that the model with explanation
corrections is steadily better in terms of explanation quality---over the
test set (top) and queries (bottom)---than the
baseline without corrections.    The predictive performance can be found in the
Appendix.  These results highlight the potential benefits of
explanatory interaction for the model's quality.

\section{Conclusion}
\label{sec:conclusion}

In this paper, we argued that explaining queries is important in assessing
trust into interactive machine learners.  Within the resulting framework of
explanatory interactive learning, we
proposed \methodname, a method that pairs model-agnostic explainers and active
learners in a modular manner.  Unlike traditional active learning approaches,
\methodname{} faithfully explains its queries in an interpretable manner and
accounts for the user's corrections of the model if it is right (wrong) for the
wrong the reasons. This opens the black-box of active learning and turns it
into a cooperative learning process between the machine and the user.  The
(boundedly rational) user is computationally limited in maximizing predictive
power globally, while the machine is limited in dealing with ambiguities
contained in a dataset.  Our experimental results demonstrate that this
cooperation can improve performance.  Most importantly, a user study validated
our key assumption, namely that explanations of interactive queries can indeed
encourage (or discourages, if appropriate) trust into the model.

There are a number of interesting avenues for future work.  Other interactive
learning approaches such as coactive learning (structured
prediction)~\cite{shivaswamy2015coactive}, active imitation
learning~\cite{judah2012active}, and mixed-initiative interactive
learning~\cite{cakmak2011mixed} should be made explanatory. In particular,
explanatory variants of the recently proposed deep active learning
approaches~\cite{gal2017deep} have the potential to further improve upon their
sample complexity. Selecting queries that maximize the information of
explanations, e.g., by using \textsc{sp-lime}~\cite{ribeiro2016should}, as well
as feeding back informative counterexample only are likely to improve
performance. If the base learner is differentiable, one may consider input
gradient explanations, even multiple ones explaining queries for qualitatively
different reasons, and then feeding back corrections via selectively penalizing
their input gradients~\cite{ross2017right}.  Generally, one should develop
interaction protocols that reduce the cognitive load on the users.


\section{Acknowledgments} The authors would like to thank Antonio Vergari,
Samuel Kolb, Jessa Bekker, and Paolo Morettin for useful discussions.  ST
acknowledges the supported by the European Research Council (ERC) under the
European Union’s Horizon 2020 research and innovation programme, grant
agreement No.~[694980] `` SYNTH: Synthesising Inductive Data Models''. KK
acknowledges the support by the German Science Foundation project ``CAML:
Argumentative Machine Learning'' (KE1686/3-1) as part of the SPP 1999 (RATIO).

\bibliographystyle{unsrt}
\bibliography{nips_2017}

\newpage
\appendix
\section{Results on the Colors Problem}

\begin{figure}[!h]
    \centering
    \begin{tabular}{cc}
        \includegraphics[width=0.4\textwidth]{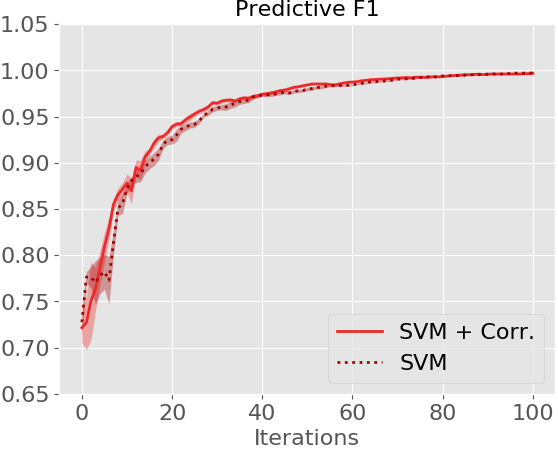}
        &
        \includegraphics[width=0.4\textwidth]{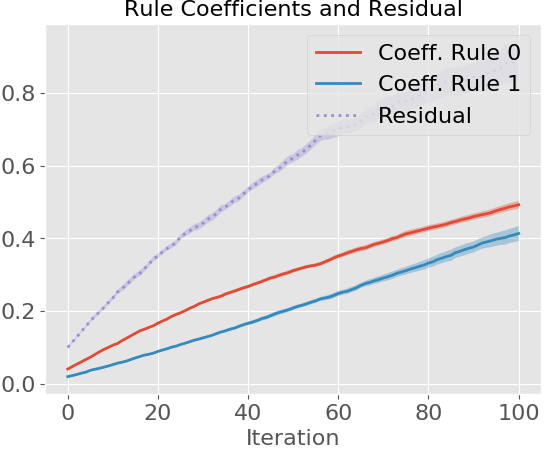}
        \\
        \includegraphics[width=0.4\textwidth]{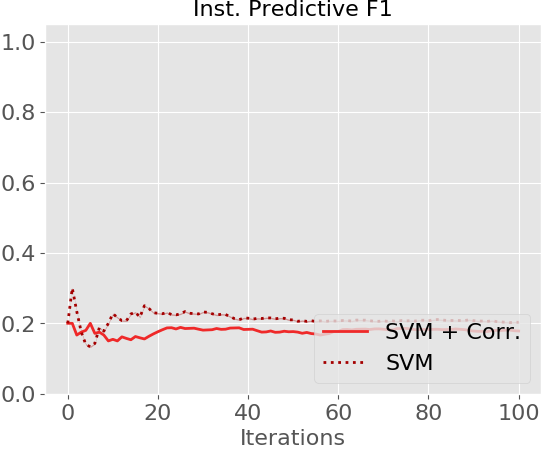}
        &
        \includegraphics[width=0.4\textwidth]{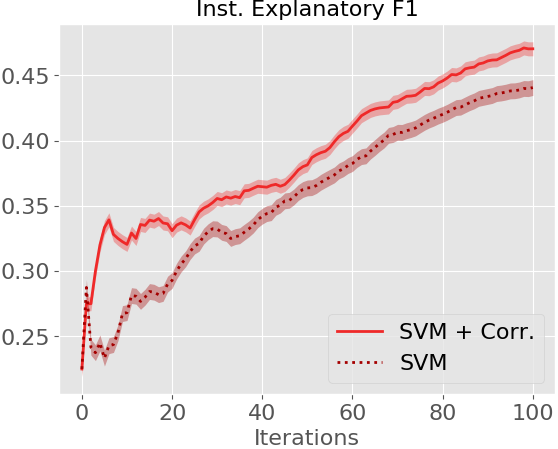}
    \end{tabular}
    \caption{\label{fig:colorsl2r0} $L_2$ SVM performance on the colors problem for rule 0. (Best viewed in color)}
\end{figure}

\begin{figure}[!h]
    \centering
    \begin{tabular}{cc}
        \includegraphics[width=0.4\textwidth]{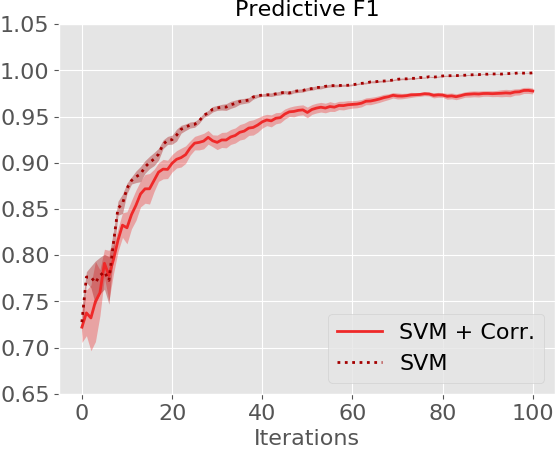}
        &
        \includegraphics[width=0.4\textwidth]{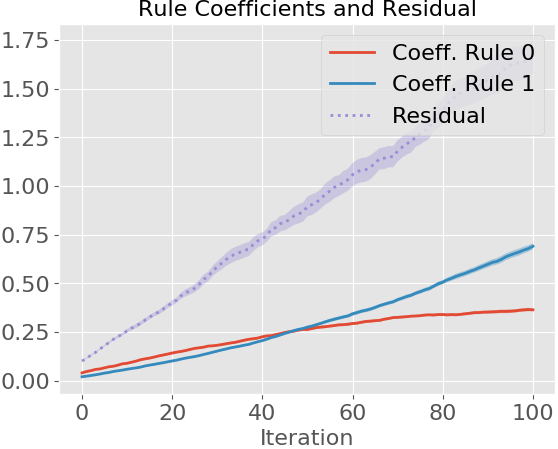}
        \\
        \includegraphics[width=0.4\textwidth]{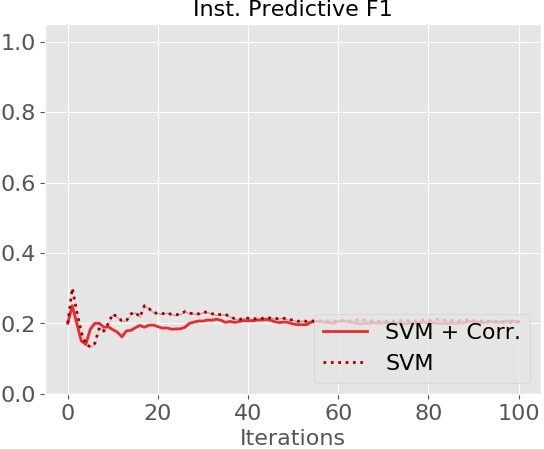}
        &
        \includegraphics[width=0.4\textwidth]{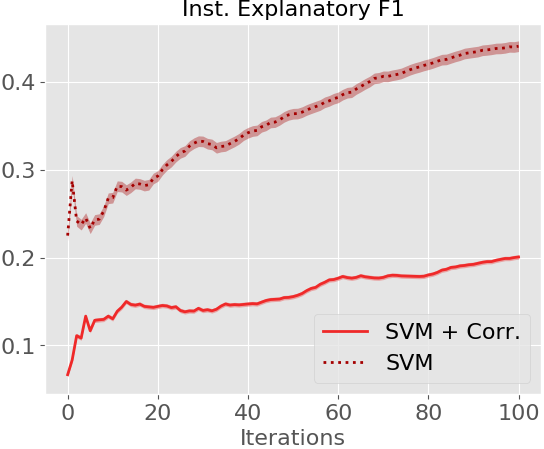}
    \end{tabular}
    \caption{\label{fig:colorsl2r1} $L_2$ SVM performance on the colors problem for rule 1. (Best viewed in color)}
\end{figure}

\begin{figure}[!h]
    \centering
    \begin{tabular}{cc}
        \includegraphics[width=0.4\textwidth]{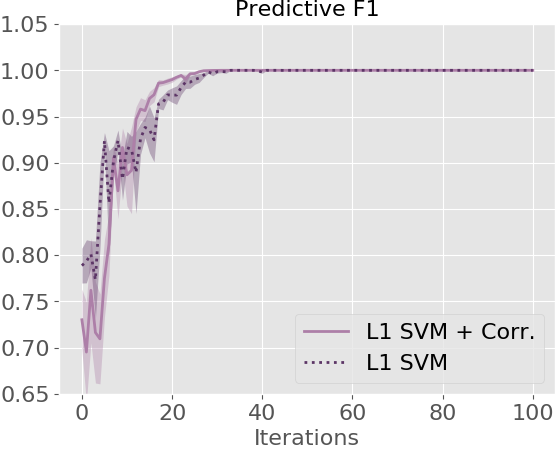}
        &
        \includegraphics[width=0.4\textwidth]{figures/colors-rule0__l1svm__least-confident__ei__coeff}
        \\
        \includegraphics[width=0.4\textwidth]{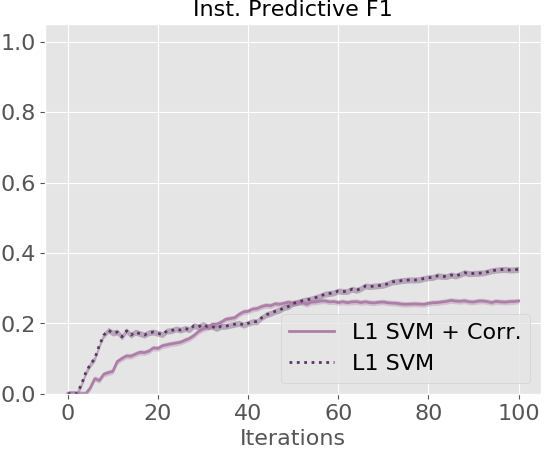}
        &
        \includegraphics[width=0.4\textwidth]{figures/colors-rule0-l1svm_instant_5}
    \end{tabular}
    \caption{\label{fig:colorsl1r0} $L_1$ SVM performance on the colors problem for rule 0. (Best viewed in color)}
\end{figure}

\begin{figure}[!h]
    \centering
    \begin{tabular}{cc}
        \includegraphics[width=0.4\textwidth]{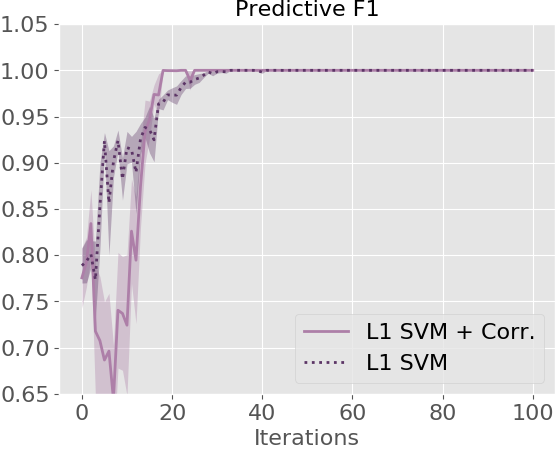}
        &
        \includegraphics[width=0.4\textwidth]{figures/colors-rule1__l1svm__least-confident__ei__coeff}
        \\
        \includegraphics[width=0.4\textwidth]{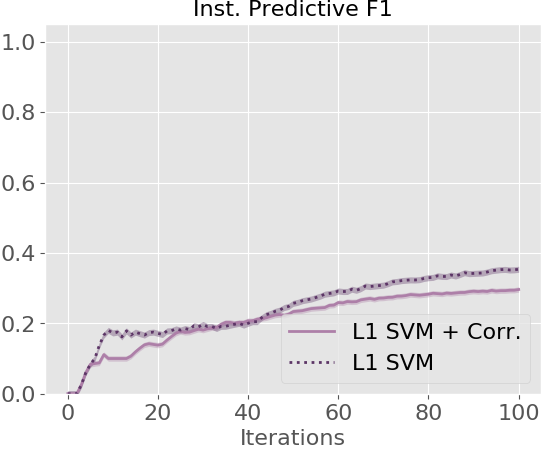}
        &
        \includegraphics[width=0.4\textwidth]{figures/colors-rule1-l1svm_instant_5}
    \end{tabular}
    \caption{\label{fig:colorsl1r1} $L_1$ SVM performance on the colors problem for rule 1. (Best viewed in color)}
\end{figure}

Here we report the complete results of \methodname{} on the colors problem.
Figure~\ref{fig:colorsl2r0} illustrates the behavior of \methodname{} for an
active $L_2$ SVM classifier when rule 0 is selected.
Figure~\ref{fig:colorsl2r1} does the same for rule 1.  All results are
$10$-fold cross-validated, the shaded areas represent the standard deviation.
The four plots represent:
\begin{itemize}

    \item Top left: the predictive $F_1$ score measured on the test set.
        Here and below, the $x$-axis represents iterations.

    \item Top right: the coefficients of the ``perfect'' weight vectors of
        two rules w.r.t. from the learned weights.

    \item Bottom left: the average instantaneous predictive $F_1$ score on the
        query instances.

    \item Bottom right: the average instantaneous cumulative explanatory $F_1$
        score on the query instances.

\end{itemize}
The results for the active $L_1$ classifier for rule 0 and 1 can be found in
Figures~\ref{fig:colorsl1r0} and \ref{fig:colorsl1r1}, respectively.

\newpage
\section{Results on the Newsgroups Dataset}

In Figure~\ref{fig:newsgroups} we report the complete results of \methodname{}
with an active logistic regression classifier applied to the ``Christian''
versus ``Atheism'' dataset.  The plots are laid out as above, except the top
right one, which in this case shows the explanatory $F_1$ performance measured
on the test set (every 20 iterations).

\begin{figure}[!h]
    \centering
    \begin{tabular}{cc}
        \includegraphics[width=0.4\textwidth]{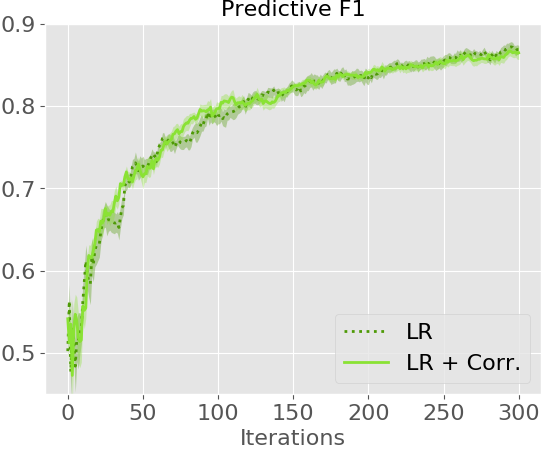}
        &
        \includegraphics[width=0.4\textwidth]{figures/newsgroups_5}
        \\
        \includegraphics[width=0.4\textwidth]{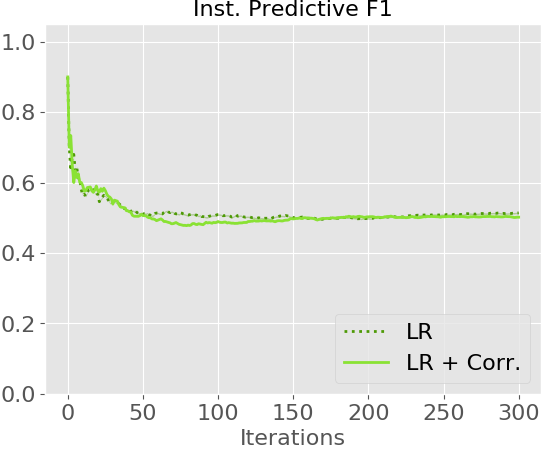}
        &
        \includegraphics[width=0.4\textwidth]{figures/newsgroups_instant_5}
    \end{tabular}
    \caption{\label{fig:newsgroups} Logistic regression on 20 newsgroups. (Best viewed in color)}
\end{figure}

\end{document}